\documentclass[11pt,a4paper]{article}
\usepackage[hyperref]{acl2021}
\usepackage{times}
\usepackage{latexsym}
\usepackage[skip=3pt]{caption}
\usepackage{graphicx}
\usepackage{booktabs}
\usepackage{url}
\usepackage{tabularx}
\usepackage{booktabs}
\usepackage{multirow}
\usepackage{xcolor}
\usepackage{latexsym}

\usepackage{listings}
\usepackage{color}
\usepackage{setspace}
\usepackage[normalem]{ulem}
\usepackage{microtype}

\definecolor{dkgreen}{rgb}{0,0.6,0}
\definecolor{gray}{rgb}{0.5,0.5,0.5}
\definecolor{mauve}{rgb}{0.58,0,0.82}
\aclfinalcopy % Uncomment this line for the final submission

\title{CoDesc: A Large Code--Description Parallel Dataset}

\author{
Masum Hasan$^1$\thanks{~~Equal contribution.}, Tanveer Muttaqueen$^1$\footnotemark[1], Abdullah Al Ishtiaq$^1$ \\
\textbf{Kazi Sajeed Mehrab}$^1$, \textbf{Md. Mahim Anjum Haque}$^1$, \textbf{Tahmid Hasan}$^1$ \\
\textbf{Wasi Uddin Ahmad}$^2$, \textbf{Anindya Iqbal}$^1$, \textbf{Rifat Shahriyar}$^1$\\ [3pt]
$^1$Bangladesh University of Engineering and Technology (BUET)\\$^2$University of California, Los Angeles\\ [3pt]
\texttt{masum@ra.cse.buet.ac.bd, 1505002.tm@ugrad.cse.buet.ac.bd}\\
\texttt{rifat@cse.buet.ac.bd}
}

\date{}

\begin{document}
\maketitle
\begin{abstract}

%Source code summarization and generation, and natural language code search depend upon parallel data between natural language and programming language. The availability of large quantities of such data has been a bottleneck for research in these fields. Although there have been many scattered attempts for these tasks, there has not been any attempt to unify and bring them under the same umbrella. 

Translation between natural language and source code can help software development by enabling developers to comprehend, ideate, search, and write computer programs in natural language.
Despite growing interest from the industry and the research community, this task is often difficult due to the lack of large standard datasets suitable for training deep neural models, standard noise removal methods, and evaluation benchmarks. This leaves researchers to collect new small-scale datasets, resulting in inconsistencies across published works.
In this study, we present CoDesc - a large parallel dataset composed of 4.2 million Java methods and natural language descriptions. With extensive analysis, we identify and remove prevailing noise patterns from the dataset. We demonstrate the proficiency of CoDesc in two complementary tasks for code--description pairs: code summarization and code search. We show that the dataset helps improve code search by up to 22\% and achieves the new state-of-the-art in code summarization. Furthermore, we show CoDesc's effectiveness in pre-training--fine-tuning setup, opening possibilities in building pretrained language models for Java. To facilitate future research, we release the dataset, a data processing tool, and a benchmark at \href{https://github.com/csebuetnlp/CoDesc}{https://github.com/csebuetnlp/CoDesc}.

\end{abstract}

\section{Introduction}
\label{sec:introduction}

%Programming is the mode of communication of programmers with computers. Being able to communicate with computers in natural language will open many new possibilities for both amateur and experienced programmers. Similar to how neural machine translation has made it possible for people of different languages to communicate conveniently \cite{bahdanau2014neural}, researchers have been trying to make the human-computer communication easier in three major aspects: natural language code search -- searching the right code snippet from a given query in natural language \cite{husain2019codesearchnet, cambronero-unif, sachdev-ncs}, source code summarization -- generating natural language description of a source code \cite{ncs, deepcom, funcom, iyer-CODENN}, and source code generation -- generating computer code from a description written in natural language \cite{wei2019code, naps}. Despite the differences in their solution approach, they all can utilize the same source code -- natural language parallel data. Although there has been notable progress in these research lines, there is no effort to combine all the available datasets such that they can be utilized for all the tasks to further push the state-of-the-art methods. 

Neural models for natural language processing have benefited from large datasets and standard evaluation benchmarks \cite{glue,superglue,rajpurkar2016squad,hermann2015teaching,common-crawl}. However, the programming language counterpart is lagging behind due to a lack in such large datasets and benchmarks.
To put this into perspective, the original Transformer network \cite{transformer} was trained on WMT'14 English--German and English--French datasets \cite{wmt-14} containing 4.5 million and 36 million parallel sentences, respectively, whereas a similar network that achieved state-of-the-art results in source code summarization has been trained on only 69 thousand code-description pairs \cite{ncs}. We argue that the existing models used for programming language tasks in the literature have a significant scope of improvement given a large, good-quality dataset, and such a dataset is the missing link for effectively applying deep learning methods on programming languages.

In this work, we collect and release a large (4.2 million) Java source code - natural language (NL) parallel dataset along with denoising methods and baseline results.
We apply our dataset to established works in both training from scratch and pre-training--fine-tuning setting and we demonstrate a notable performance gain in both settings. We gain 10\% to 22\% improvement over baseline code search models using CoDesc, and attain performances comparable to models having 8$\times$ more parameters. We achieve a new state-of-the-art BLEU score of 45.89 in code summarization by pretraining a Transformer network with our dataset for two epochs. With extensive empirical analysis, we propose a set of noise removal techniques for the source code and the NL descriptions in our dataset.

% Our main contribution in this paper is a large (4.2 million) source code - natural language parallel dataset with denoising methods and baseline results.
Our work brings together several datasets and multiple tasks on the intersection of Natural Language Processing (NLP) and Software Engineering (SE), such as code summarization, code search and code synthesis, and allows researchers to compare their methods on the same benchmark. It also opens the door for building large pretrained models to jointly learn code and NL representations that can be leveraged in downstream tasks  that do not have adequate data, such as, code refactoring, clone detection, etc. as done by \citet{codebert}.

% and also natural language aided code repair \cite{review4repair}, automatic comment generation, code paraphrasing, etc. 

% Potential applications
% - Building pretrained language models for source code
% - Code summarization
% - Code search
% - Code generation/synthesis

% - Code similarity
% - Code completion
% - Code paraphrasing
% - Clone detection
% - Natural Lanugage aided code repair
% - Automatic comment generation
% - Code refactoring
% - Automatic documentation generation
% - Building new architectures for source code

% Table
\begin{table*}
\resizebox{\linewidth}{!}{
% \setlength\arrayrulewidth{.7pt}
% \begin{tabular}{l|r|r|r|r|r|r|r|r|r}
\begin{tabular}{l|c|c|c|c|c|c|c|c|c}
\hline
\multicolumn{1}{l|}{\multirow{3}{*}{\textbf{Source}}} & \multicolumn{1}{c|}{\multirow{3}{*}{\textbf{\begin{tabular}[c]{@{}c@{}}\#Projects\end{tabular}}}} & \multicolumn{1}{c|}{\multirow{3}{*}{\textbf{\begin{tabular}[c]{@{}c@{}}\#Raw\\ data\end{tabular}}}} & \multicolumn{1}{c|}{\multirow{3}{*}{\textbf{\begin{tabular}[c]{@{}c@{}}\#Clean\\ data\end{tabular}}}} & \multicolumn{3}{c|}{\textbf{Code}} & \multicolumn{3}{c}{\textbf{Description}} \\ \cline{5-10} 
\multicolumn{1}{c|}{} & \multicolumn{1}{c|}{} & \multicolumn{1}{c|}{} & \multicolumn{1}{c|}{} & \multicolumn{1}{c|}{\textbf{\begin{tabular}[c]{@{}c@{}}\#Unique\\ tokens\end{tabular}}} & \multicolumn{1}{c|}{\textbf{\begin{tabular}[c]{@{}c@{}}Avg \\ len\end{tabular}}} & \multicolumn{1}{c|}{\textbf{$\leq$200}} & \multicolumn{1}{c|}{\textbf{\begin{tabular}[c]{@{}c@{}}\#Unique \\ tokens\end{tabular}}} & \multicolumn{1}{c|}{\textbf{\begin{tabular}[c]{@{}c@{}}Avg \\ len\end{tabular}}} & \multicolumn{1}{c}{\textbf{$\leq$50}} \\ \hline
CSN-Java & N/A & 542,991 & 490,169 & 284,214 & 140.41 & 83.42 & 168,507 & 25.14 & 89.42 \\
DeepCom & 9,714 & 588,108 & 424,028 & 306,422 & 128.35 & 84.04 & 91,933 & 17.80 & 94.76 \\
FunCom & 28,000 & 2,149,121 & 2,130,247 & 469,354 & 51.30 & 99.83 & 399,338 & 15.52 & 95.87 \\
CONCODE & 33,000 & 2,184,310 & 733,040 & 131,852 & 33.75  & 99.99 & 166,239 & 14.87 & 96.27 \\
CSN-Py2Java & N/A & 456,000 & 434,032 & 414,018 & 163.78 & 72.32 & 223,277 & 57.11 & 68.69 \\
CoDesc (All) & N/A & 5,920,530 & 4,211,516 & 1,128,909 & 77.97 & 93.53 & 813,078 & 21.04 & 92.28 \\
\hline
\multicolumn{10}{l}{Balanced \textit{train-valid-test} split for CoDesc data} \\
\hline
\textit{train} & \multicolumn{1}{c|}{-} & \multicolumn{1}{c|}{-} & 3,369,218 & 991,395 & 78.01 & 93.53 & 718,204 & 21.05 & 92.28 \\
\textit{valid} & \multicolumn{1}{c|}{-} & \multicolumn{1}{c|}{-} & 421,149 & 269,435 & 77.73 & 93.51 & 188,145 & 21.08 & 92.26 \\
\textit{test} & \multicolumn{1}{c|}{-} & \multicolumn{1}{c|}{-} & 421,149 & 269,318 & 77.88 & 93.55  & 187,230 & 20.97 & 92.33 \\
% \multicolumn{11}{l}{Code Summarization training data} \\ 
% \hline
% \textit{train-small} & & \multicolumn{1}{c|}{-} & 69707 & 26328 & 122.95 & 84.21 & 15066 & 18.18 & 95.21 \\
% \textit{CoDesc-train} & & \multicolumn{1}{c|}{-} & 4175165 & 44127 & 77.11 & 93.78 & 24537 & 21.34 & 91.98 \\
% \textit{validation} & & \multicolumn{1}{c|}{-} & 8714 & 14148 & 122.52 & 84.41 & 8100 & 18.33 & 95.38 \\
% \textit{test} & & \multicolumn{1}{c|}{-} & 8714 & 13944 & 124.19 & 84.09 & 7939 & 18.07 & 95.23 \\
\hline 
\end{tabular}
}
\caption{Statistics of CoDesc datasets and a balanced train-valid-test split. $\leq$200 and $\leq$50 indicates the percentage (\%) of data where source code and description are smaller than 200 and 50 tokens, respectively.}
\label{tab:data-description}
% \vspace{-0.45cm}
\end{table*}

\section{Related Works}
\label{sec:related-works}
\paragraph{Code-Description Parallel Datasets}

% {\color{blue} @Masum, please rewrite the related works section. Organize this section in 2 paragraphs. 
% First paragraph talking about datasets and the 
% second paragraph discusses prior work on code search / summarization. -- Wasi}

With the advent of data-driven code search and code summarization methods, several datasets are proposed to facilitate research in code-NL parallel tasks.
%\cite{gu-codenn, sachdev-ncs, cambronero-unif, husain2019codesearchnet}. 
\citet{husain2019codesearchnet} introduced \textsc{CodeSearchNet (CSN)}, a benchmark for code search techniques with 2.1 million code-NL parallel data in 6 programming languages, 6.4 million monolingual code data, a leader-board, and baseline results with 5 code search techniques. CONCODE \cite{concode}, DEEPCOM \cite{deepcom}, FunCom \cite{funcom} are some notable dataset papers that released 2.18 million, 2.15 million, and 0.59 million parallel data respectively. \citet{pymt5} released a parallel corpus of 26 million monolingual Python methods and 7.7 million method-docstring pairs. CoNaLa \cite{conala} is a Python line by line natural language description dataset containing nearly 3k parallel data.

\paragraph{Code Search and Summarization}
CODE-NN \cite{iyer-CODENN} is a pioneering work in data-driven code summarization.
The CodeSearchNet dataset paved the way for CodeBERT \cite{codebert}, a pretrained BERT \cite{bert} model trained on CSN data with Masked Language Modeling (MLM) \cite{bert} and Replaced Token Detection (RTD) \cite{clark2020electra} objective that achieved a high performance in the CSN benchmark.  \citet{wei2019code} proposed a dual learning method that simultaneously trained code summarization and code generation and improved both of them using 60k parallel data. In the same dataset, \citet{ncs} achieved state-of-the-art results in source code summarization using a Transformer network \cite{transformer}. Along with the mentioned dataset, \citet{pymt5} presented PyMT5, a text to text transformer that notably improved method generation and code summarization. \citet{plbart} collected more than 300 GB monolingual code and NL data, and trained PLBART, a pretrained seq2seq model for both program understanding and comprehension. 

%\citet{guo2020graphcodebert} presented GraphCodeBERT, a subsequent work that improve the results of CodeBERT by utilizing the graph structure in source code.

\section{CoDesc Dataset}
\label{sec:codesc}
% \lstinputlisting[language=Java]{Images/file.java}

\subsection{Data Sources}
\label{subsec:data-sources}

We collect our data from several sources and formulate rules for data cleaning. 5 of the authors spent 45-50 man-hours manually going over the dataset to identify patterns of noises in different data sources. Upon group discussion, common patterns were identified and a noise removal method was established. Details about these noise patterns are provided in Appendix A.

One of the datasets used in CoDesc is \textsc{CodeSearchNet (CSN)}\footnote{\href{https://github.com/github/CodeSearchNet}{https://github.com/github/CodeSearchNet}} \cite{husain2019codesearchnet} - a parallel method-description dataset for code search. Furthermore, other datasets used are DeepCom\footnote{\href{https://github.com/xing-hu/DeepCom}{https://github.com/xing-hu/DeepCom}} \cite{deepcom}, CONCODE\footnote{\href{https://github.com/sriniiyer/concode}{https://github.com/sriniiyer/concode}} \cite{concode}, FunCom\footnote{\href{http://leclair.tech/data/funcom/}{http://leclair.tech/data/funcom/}} \cite{funcom} - datasets created for code summarization. The \textsc{CodeSearchNet} dataset originally contained 6 programming languages, from which the Java methods are directly used in CoDesc, however, the Python methods are used after being automatically translated to Java. We combine all aforementioned datasets to create CoDesc. Appendix B shows a sample code-description parallel data from each of these datasets. Table \ref{tab:data-description} describes our data sources and their characteristics in detail.

% More error analysis: https://docs.google.com/document/d/100XZ91XzETLqYAWgb1iOrYPFAlFJFMfZeKw2xXSYEPA/edit
% https://docs.google.com/document/d/1fL3Ta6v_S-hv0TYBsKDe4ug5YG2WJ-0SMtgkJQCczhA/edit

% \textbf{\textsc{CSN} Python to Java Translation.} 
\paragraph{\textsc{CSN} Python to Java Translation} 
To utilize maximum possible data from the \textsc{CSN Corpus}, we translate the Python methods to Java using TransCoder \cite{transcoder}, a state-of-the-art, neural source-to-source compiler.
We modified and re-released the open-source implementation of TransCoder\footnote{\href{https://github.com/csebuetnlp/TransCoder}{https://github.com/csebuetnlp/TransCoder}}, enabling it to translate data in batches instead of one at a time, and resulting in a 16X faster translation. 
%Translating the dataset took approximately 96 hours of compute time.
Upon empirical inspection, we found that the converted Java codes are human-readable and bear a strong resemblance to the original Python code intent. The converted codes seem correct to the human eye and their syntax matches with Java syntax. However, a few cases the transcompiler suffers are -- converting to Java library methods, and converting from Python coding conventions that does not have a Java equivalent (e.g. use of \textsc{self}). These conversion errors, however, were not severe enough to affect our model to learn the NL-source code mapping.

\subsection{Data Cleaning and Noise Removal}
\label{subsec:clean-data}

We created an easy-to-use, parameterized data processing tool for removing the different types of noise that we observed in our dataset. From the natural language descriptions, we remove symbols and characters that do not carry a meaning in a natural language description, such as, comment tags (e.g., \texttt{//, /*, */}), stray code characters (e.g., \texttt{@, \#, \{, \},} etc.), HTML and XML tags,  non-ASCII and escape characters, and some patterns of autogenerated tags (e.g., \texttt{@param, @return, @throws,} etc.).
% @link, @code, @inheritdoc,
From source code, we remove comments and the non-ASCII and escape characters. In previous studies, many meaningful data are discarded due to having some noisy patterns/symbols either in the code or description \cite{husain2019codesearchnet,concode,funcom}. We identify and remove the noisy part of the data points without excluding them from the dataset to reduce data loss during preprocessing.
% This helps us reduce data loss in preprocessing.

For both source code and NL description, we split CamelCase and snake\_case code tokens into subtokens (e.g., Camel Case, snake case) and separate linked alphabets and numbers (e.g., var0 to var 0) \cite{ncs, funcom}. After the aforementioned processing, we remove the data points where the source code is less than 3 tokens, or the description contains less than 2 alphabets \cite{husain2019codesearchnet}. We lowercase the natural language as the case is not necessary for describing codes. We release our data processing tool along with the CoDesc dataset for applying the dataset to diverse tasks.

\subsection{Dataset Characteristics}
% The dataset we release consists of Java methods and natural language (NL) descriptions both before and after the processing described in Section \ref{subsec:clean-data}, along with the source-dataset name of each datapoint.
After the previous steps, we are left with nearly 4.2 million Java method and description parallel data.
Table \ref{tab:data-description} presents the statistics characteristics of our dataset. The combined CoDesc dataset consists of more than one million unique tokens, which is significantly larger than natural language vocabulary \cite{nlp-vocab}. This can be partially attributed to inseparable multi-words (e.g. `updateproductvariationlocalizeddeltaprice') in our dataset. Hence, we perform BPE \cite{bpe} tokenization in our preprocessing pipeline. We also see that although the average token length of Java source codes vary in the different dataset sources, the natural language descriptions have a relatively uniform length. We create a balanced, deduplicated, and representative \textit{train--valid--test} dataset by splitting individual source-dataset in 8:1:1 ratio (Table \ref{tab:data-description}).

\section{Experiments}
\label{sec:experiments}

We evaluate our code-description corpus in two well-known complementary tasks: source code summarization and natural language code search. In this section, we demonstrate that models trained on CoDesc bring about a noticeable improvement over two established baselines in code search and code summarization. Each benchmarking follows a standard cleaning, preprocessing, and train-test de-duplication process.

% Each baseline task contains a separate tokenization scheme, and validation and test data. Considering the magnitude of the CoDesc dataset, it is likely that some of the test and validation data will be present in the CoDesc. Hence, prior to both Code Search and Code Summarization, all data is tokenized and formatted in the corresponding scheme, and repetition with test or validation data is removed from the train set.

\subsection{Natural Language Code Search}

\begin{figure}
    \centering
    \includegraphics[width=\linewidth]{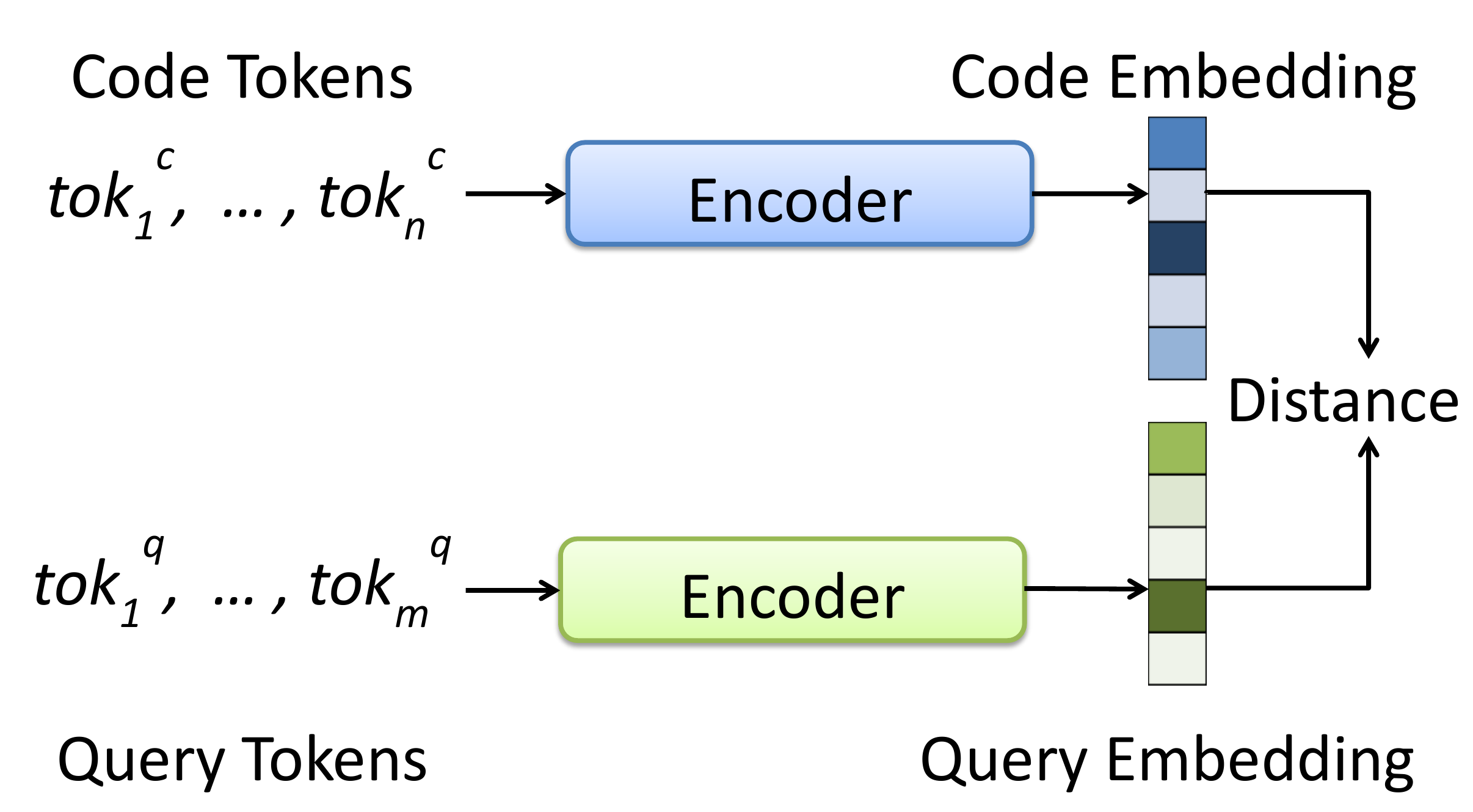}
    \caption{Code search model architecture; code and NL (query) encoders jointly train to reduce their embedded distance. During search, we select the code that is closest to the query in their shared embedding space.}
    \vspace{1mm}
    \label{fig:search-model}
\end{figure}

We use the code search models used by \citet{husain2019codesearchnet} that jointly trains a source code and an NL encoder networks to minimize their encoded vector distance (Figure. \ref{fig:search-model}).
We apply our dataset on the \textsc{CodeSearchNet (CSN)}  \cite{husain2019codesearchnet} -- a well-studied benchmark in the semantic code search literature. We train 5 different encoder networks (Table. \ref{tab:csn-results}) with the CSN Java dataset, and CoDesc respectively. We compare our results with CodeBERT and RoBERTa (code) \cite{codebert}, two pretrained models achieving state-of-the-art results in CSN Benchmark. They are trained with a Masked Language Modeling (MLM) \cite{bert} objective on 2.1 million bimodal code-NL data, and 6.4 million unimodal data released with \textsc{CodeSearchNet}.

% \textbf{Results.} 
\paragraph{Results} 
We use Mean Reciprocal Rank (MRR) -- the commonly used  evaluation metric for code search \cite{husain2019codesearchnet,sachdev-ncs,cambronero-unif} as the evaluation criteria for code search. Table \ref{tab:csn-results} shows our results, along with state-of-the-art models \cite{liu2019roberta, codebert} that have nearly 8-10 times more parameters than the baseline networks and a more complex training objective. We achieve remarkably close performance with the state-of-the-art models with much simpler and smaller networks.

\begin{table}
\setlength\arrayrulewidth{.75pt}
\resizebox{\columnwidth}{!}{
\begin{tabular}{l|c|c|c}
\hline
\multicolumn{1}{c|}{\multirow{2}{*}{\textbf{Model}}} & \multirow{2}{*}{\textbf{\#Param}} &  \multicolumn{2}{c}{\textbf{CSN test MRR}}\\ \cline{3-4} 
& & \multicolumn{1}{c|}{\textbf{CSN-Java}} & \multicolumn{1}{c}{\textbf{CoDesc}} \\
\hline
NBOW & 11.6 M & 0.589 & 0.683 \\
RNN & 12.6 M & 0.582 & 0.679 \\
Sel-attn & 13.6 M & 0.583 & 0.723 \\
1D Conv & 16.4 M & 0.520 & 0.686 \\
Conv self-attn & 16.0 M & 0.509 & \textbf{0.729} \\
\hline
\multicolumn{4}{l}{State-of-the-art models} \\
\hline
RoBERTa (code) & 125 M & \multicolumn{2}{c}{0.721} \\
CodeBERT & 125 M & \multicolumn{2}{c}{\textbf{0.748}} \\
%GraphCodeBERT & 125 M & \multicolumn{2}{c}{\textbf{0.757}} \\
\hline
\end{tabular}
}
\caption{Baseline models trained with default dataset and CoDesc, along with, comparison with SoTA pretrained models in \textsc{CSN} Java test set. Training on CoDesc outperforms training on CSN-Java only, and it is comparable to SOTA with 8x fewer parameters.}
\label{tab:csn-results}
% \vspace{-2mm}
\end{table}

\subsection{Source Code Summarization}
For this task, we follow the methodology proposed by \citet{ncs}. They used a seq2seq Transformer \cite{transformer} network with 77M parameters with relative positional encoding \cite{shaw} and copying mechanism \cite{copy} and achieved state-of-the-art results.

% \textbf{Data preparation.} 
\paragraph{Data preparation} 
\citet{ncs} used a Java dataset released by \citet{hu} and preprocessed by \citet{wei2019code} consisting of \textit{training}, \textit{validation}, and \textit{test} datasets of size 69,708, 8,714, and 8,714 respectively. We refer to this training data as \textit{train-small}. We create a new dataset \textit{CoDesc-train} by combining \textit{train-small} with CoDesc.
We replace all literals as \citet{wei2019code} and tokenize the dataset using Character BPE Tokenization \cite{bpe} to create the same size vocabulary as the previous works.
%\citet{ncs} and \citet{wei2019code}.

\begin{table}[t]
\centering
\setlength\arrayrulewidth{.6pt}
\resizebox{\columnwidth}{!}{
\begin{tabular}{@{}l|c@{\hskip 0.1in} c@{\hskip 0.1in} c@{}}
\hline
\textbf{Methods} & \textbf{BLEU} & \textbf{METEOR} & \textbf{ROUGE-L} \\ \hline
% CODE-NN \cite{iyer-CODENN} & 27.6 & 12.61 & 41.1 \\
% Tree2Seq \cite{tree2seq} & 37.88 & 22.55 & 51.5 \\
% RL+Hybrid2Seq \cite{rl-hybrid} & 38.22 & 22.75 & 51.91 \\
% DeepCom \cite{deepcom} & 39.75 & 23.06 & 52.67 \\
% API+CODE \cite{hu} & 41.31 & 23.73 & 52.25 \\
% Dual Model \cite{wei2019code} & 42.39 & 25.77 & 53.61 \\
% \hline
Transformer & 44.58 & 26.43 & 54.76 \\ % reported
%Transformer \cite{ncs} & 44.76 & 27.13 & 54.81 \\ % reproduced
%Transformer (200 tokens input) & 45.07 & 27.15 & 55.31 \\
CoDesc pretrained & \textbf{45.89} & \textbf{28.01} & \textbf{56.59} \\ 
\hline
\end{tabular}
}
\caption{Code summarization with \citet{ncs} proposed Transformer network with and without pretraining with CoDesc.}
\label{tab:ncs-results}
% \vspace{-1mm}
\end{table}

\begin{figure*}
% \resizebox{.5\textwidth}{!}{
\begin{lstlisting}
public void makeImmutable(){ 
	if(mutable){ 
		if(results ! = null){ 
			int length = results.size(); 
			for(int i = NUM; i < length; i + +){ 
				Result result = (Result)results.get(i); result.makeImmutable(); 
			} results = Collections.unmodifiableList(results); 
		} mutable = BOOL; } }
\end{lstlisting}
\vspace{-4pt}
{\small
\renewcommand{\baselinestretch}{.4}
\textcolor{violet}{Human written:} makes the object immutable \\
\textcolor{orange}{Transformer prediction} (BLEU: 1.0): makes the object immutable \\
% \hline
\uline{\textcolor{cyan}{CoDesc pretrained model prediction} (BLEU: 0.12): if there are any object in the list then the object is not immutable \hfill}
% \\ \vspace{-5pt}
% \rule{\linewidth}{0.4pt}
}
% \vspace{-5pt}
% \noindent\makebox[\linewidth]{\rule{\linewidth}{0.4pt}}
% \singlespacing
% \vspace{3pt} %7
\caption{CoDesc pretrained model generates more descriptive summary, even in cases it achieves lower score.}
% \vspace{-1mm}
\label{fig:sample-data}
\end{figure*}

% Redo this paragraph. Describe NCS work first, then your changes.
% \textbf{Training.} 
\paragraph{Training} 
We train a Transformer model proposed by \citet{ncs} with \textit{CoDesc-train} dataset. We use Adam optimizer with an initial learning rate of $10^{-4}$, mini-batch size of 32, and dropout rate 0.2, vocabulary size 50k for code and 30k for NL. However, we use maximum input length of 200 token instead of 150 based on our observation of CoDesc dataset from Table \ref{tab:data-description}. Each epoch of the model took nearly 8 hours in an NVIDIA V100 16GB GPU. In comparison, the \textit{train-small} dataset took 8.5 minutes only. For limitation of computational resource, we saved the network weights after training it with the large dataset for two epochs, and to be consistent with the original implementation, trained them further with the \textit{train-small} dataset for a maximum of 198 more epochs. We perform an early stop if the validation performance does not improve for consecutive 20 epoch. The pretraining provides the network parameters a more favorable initialization than random, helping the network find better local minima.

% \textbf{Results.} 
\paragraph{Results} 
Table \ref{tab:ncs-results} shows that our two epoch pretraining with CoDesc significantly improves the state-of-the-art code summarization methods in all three evaluation metrics -- BLEU \cite{bleu}, METEOR \cite{meteor}, and ROUGE-L \cite{rouge}. We observe that the pretrained model often generates more descriptive summary even when it achieves lower BLEU score (Fig. \ref{fig:sample-data}).
We believe the model has more room for improvement with further pretraining and we wish to validate this in future work.

% \vspace{-15pt}
\subsection{Ablation \& Analysis}

To quantify the effect of individual data sources and our noise removal methodology, we train each dataset before and after applying our data cleaning method using an NBOW model and test them in the CSN benchmark using their released test set.

Although our collected data was already cleaned by the respective authors, Table \ref{tab:ablation} shows that the performance of every dataset improves drastically after our noise removal. Interestingly, without our extra layer of data cleaning, CoDesc dataset performs worse than training with only CSN data although being significantly larger. This shows the importance of a standard cleaning and processing method. Moreover, CSN (Java) have the highest accuracy, which can be attributed to the fact that it came from the same distribution of data as the evaluation and test sets, and hence contains similar tokens and patterns \cite{husain2019codesearchnet}. We can see from Table \ref{tab:ablation} that the model trained with CSN (Python2Java) achieves an MRR score of 0.5548. Although this score is lower than other datasets, it is still a good indication that the translated data is helping the model is to learn NL-code association.

\begin{table}
\resizebox{\columnwidth}{!}{
\setlength\arrayrulewidth{.7pt}
\begin{tabular}{l|ccr}
\hline
\textbf{Dataset} & \multicolumn{1}{c}{\textbf{\begin{tabular}[c]{@{}c@{}}Raw\\ data\end{tabular}}} & \multicolumn{1}{c}{\textbf{\begin{tabular}[c]{@{}c@{}}Clean\\ data\end{tabular}}} & \multicolumn{1}{c}{\textbf{Inc. (\%)}} \\ 
\hline
CSN (Java) & 0.5870 & 0.6427 & 5.57 \\
DeepCom & 0.4677 & 0.6069 & 13.92 \\
FunCom & 0.5379 & 0.6366 & 9.87 \\
CONCODE & 0.5444 & 0.6234 & 7.90 \\
\textsc{CSN} (Python2Java) & 0.5081 & 0.5546 & 4.65 \\ 
CoDesc (All) & 0.5852 & 0.6826 & 9.74 \\
\hline
\end{tabular}
}
\caption{MRR of individual datasets (Section \ref{subsec:data-sources}) before and after noise removal.}
\label{tab:ablation}
\vspace{-2mm}
\end{table}

\paragraph{New Benchmark Results in Code Search}
We provide a new set of benchmark results for CoDesc dataset in natural language code search. We train, validate, and test an NBOW, an RNN, and a Self-attn code search network with the balanced \textit{train}, \textit{validation}, and \textit{test} data shown in Table \ref{tab:data-description}. The three models achieve MRR score of \textbf{0.812}, \textbf{0.766}, and \textbf{0.839} respectively.

\section{Discussion and Conclusion}
\label{sec:conclusion}

In this work, we have accumulated CoDesc -- a large code-description parallel dataset and established baseline results.
CoDesc brings a noteworthy improvement in two tasks: code search and code summarization. We believe CoDesc will serve as a base for future studies on code-description joint tasks.
% We also demonstrate that pretraining with CoDesc can improve downstream tasks. This opens the door for creating large, pretrained models with CoDesc that can be utilized for downstream tasks that have limited datasets.
We also show that automatically translated source code from a source-to-source compiler can be applied in a code-NL parallel task, suggesting that, translating our Java dataset to other programming languages can also be helpful.

The most striking finding of our study is that, by training with 2X larger parallel data, we achieve equivalent performance to models having 8X parameters \cite{codebert} in code search. This raises an interesting question: 
% should the focus of code--description studies be on larger networks, or should it be on larger datasets?
are we fully utilizing the model capacities in code--description studies?
From our pretraining results in code summarization, it can be reasonably assumed that pretraining with our large dataset the larger models will also improve further.
% In future studies, we would like to train large, pretrained language models, similar to \citet{codebert}, for source code - description tasks. 
% Given that the focus of this study is to propose and validate the CoDesc dataset, we refrain from proposing new methodologies. 
In future works, we wish to apply new techniques for code search, code summarization, along with exploring our dataset for general-purpose code synthesis, where the best models are still struggling in accuracy \cite{wei2019code,codegen2}.

\section*{Acknowledgement}
\label{sec:acknowledgement}
We thank the ICT Division, Bangladesh for funding the project and Intelligent Machines Limited for providing the cloud support.

\bibliographystyle{aclnatbib}
\bibliography{ref}

\clearpage
\appendix

\twocolumn[{%
 \centering
 \Large\bf Supplementary Material: Appendices \\ [20pt]
}]
\section{Dataset Details}

% \textbf{\textsc{CodeSearchNet (CSN) Corpus}.}
\paragraph{CodeSearchNet (CSN) Corpus} is a code search dataset for 6 programming languages
\cite{husain2019codesearchnet}.\footnote{\href{https://github.com/github/CodeSearchNet}{https://github.com/github/CodeSearchNet}}
Despite the authors' effort for data cleaning, in our observation, \textsc{CSN Corpus} is one of the noisiest. The dataset contains duplicate descriptions, inseparable multi-words (e.g., \texttt{updateproductvariationlocalizedde- ltaprice, updatelocationinventory}), XML tags (e.g., \texttt{<tt>, <soup>, <sub>}),
% <pre>, <p>
non-English documentation, non-ASCII and escape characters, unwanted symbols (e.g., \texttt{@, \#, \{, \}}), deprecated methods and descriptions, comments inside code, annotations (e.g., \texttt{@link, @code, @inheritdoc}) in description, etc. Datapoints truncated by TransCoder during Python to Java translation (total 27,471) are marked with a special flag in our released corpus. 

%They collected a total of 6,452,226 functions in 6 programming languages: Go, Java, JavaScript, PHP, Python, Ruby. Among them, 2,326,976 functions contained natural language documentations.% We collected the Java and Python dataset from their data. Using a state-of-the-art program translation method, TransCoder \cite{transcoder}, we translate the Python source code to Java. %, and we collect the  1,046,493 functions with their corresponding documentation to our dataset. %Our program translation method is described in the next section. 

% \textbf{DeepCom.} 
\paragraph{DeepCom}
\citet{deepcom} released a dataset of 588,108 Java method and documentation pairs collected from 9,714 GitHub projects for code summarization\footnote{\href{https://github.com/xing-hu/DeepCom}{https://github.com/xing-hu/DeepCom}}. Similar to \textsc{CodeSearchNet} \cite{husain2019codesearchnet}, they considered the first sentence of a documentation as the summary of the method as it typically describes the functionalities of Java methods. They filter out empty and single world descriptions and the setter, getter, constructors, and test methods, since they are easy for a model to summarize. In our manual analysis, we found HTML tags (e.g. \texttt{<tt>, <p>, <p class = …> , <li>, <ul>}), comment tags, annotations, escape characters inside descriptions, empty parentheses as descriptions, repetitive and non-meaningful descriptions, comments inside source code, etc. Despite the authors' claim, we found numerous test methods in the dataset, which were mostly meaningful data.

% \textbf{CONCODE. }
\paragraph{CONCODE} \citet{concode} released a dataset named CONCODE, collected by mining nearly 33,00 GitHub repositories\footnote{\href{https://github.com/sriniiyer/concode}{https://github.com/sriniiyer/concode}}. In their preprocessed dataset, they replaced the names of the identifier and method names with generic terms, (e.g., \texttt{arg0, loc0, function,} etc.) and replaced all string literals with constants. This created a discrepancy with the other datasets, hence, we opted for their unprocessed dataset rather than the processed version. The unprocessed dataset released with CONCODE contained approximately 2.1 million Java methods and lowercased Javadoc-style document pairs. Upon duplicate removal, we were left with 733,878 datapoints. 

Although some noises were present in this dataset, we found this data to be least noisy in manual observation. We find that because of lower casing the documentations, some CamelCase tokens became inseparable. The dataset also contained non-English comments with English alphabets (mostly Italian). We found these documents hard to identify and remove.

% \textbf{FunCom.} 
\paragraph{FunCom}
\citet{funcom} released a dataset of over 2.1 million pairs of Java methods and one-sentence method descriptions from over 28k Java projects\footnote{\href{http://leclair.tech/data/funcom/}{http://leclair.tech/data/funcom/}}. They collected this dataset by filtering over 51 million Java methods from UCI Source Code datasets \citep{uci-data}. In their preprocessing step, \citet{funcom} removed all datapoints where the method is more than 100 tokens long, or the method description is over 13 tokens or below 3 tokens.

In our observation of this dataset, we found method descriptions containing HTML tokens (e.g. \texttt{<tt>}, annotations (e.g., \texttt{@link, @param}), comment tokens, unwanted symbols, solely non-alphabetic characters, etc. It also contained comments inside methods, and a large portion of the data were \texttt{getter}, \texttt{setter}, \texttt{tester}, and \texttt{toString} methods.

\section{Sample Data}

% \begin{figure*}
% 0

% \vspace{-15pt}
% \vspace{-15pt}
\begin{lstlisting}
protected void hideTabs(){
  if (getPageCount() <= 1) {
    setPageText(0,"");
    if (getContainer() instanceof CTabFolder) {
      ((CTabFolder)getContainer())
      .setTabHeight(1);
      Point point = getContainer().getSize();
      getContainer().setSize(point.x,point.y + 6);
    }
  }
}
\end{lstlisting}
\vspace{-4pt}
{\small
% \halfspacing
\textcolor{violet}{Description:} if there is just one page in the multi - page editor part , this hides the single tab at the bottom. (DeepCom) \\ \vspace{-5pt}
%\hline
% 5
\singlespacing
}

% \clearpage

\begin{lstlisting}
@Exported
    public boolean isIdle() {
        lock.readLock().lock();
        try {
            return workUnit == null && executable == null;
        } finally {
            lock.readLock().unlock();
        }
    }
\end{lstlisting}
\vspace{-4pt}
{\small
%\halfspacing
\textcolor{violet}{Description:} returns true if this executor is ready for action. (CodeSearchNet) \\ %\vspace{-5pt}

% 1
\singlespacing
}

% \vspace{-15pt}
\begin{lstlisting}
public static void dbCommand(ParserArgs args){
  final Synergy synergy =(Synergy)args.get("synergy");
  if(synergy.reset){
    synergy.resetDb();
    synergy.update = true;
  }
  if(synergy.update){
    synergy.updateDb();
  }
}
\end{lstlisting}
\vspace{-4pt}
{\small
%\halfspacing
\textcolor{violet}{Description:} manages synergy db state (CodeSearchNet- Python to Java) \\ %\vspace{-5pt}
% 2
\singlespacing
}

\begin{lstlisting}
Object getBean(String beanName){ 
	if(null == beanName){ 
		return null; 
	} 
	return applicationContext.getBean(beanName); 
}
\end{lstlisting}
\vspace{-4pt}
{\small
% \halfspacing
\textcolor{violet}{Description:} this method is used to retrieve a bean by its name. note that this may result in new bean creation if the scope is set to ``prototype" in the bean configuration file. (CONCODE) 
% \vspace{-5pt}
% 3
\singlespacing
}

% \vspace{-15pt}
\begin{lstlisting}
public void sort (boolean transformChanged) { 
	if (list Size > 1){ 
		if (tlist == null || tlist.length != list.length){
			tlist = list.clone(); 
		} else { 
			System.arraycopy(list, 0, tlist, 0, list.length); 
		} 
		if (transform Changed) { 
    		for(int i = 0; i < listSize; i++) { 
    		    list[i]
    		    .computeLastDistance(owner); 
    		}
		} 
		SortUtil.msort(tlist, list, 0, list Size - 1, c); 
	} 
}
\end{lstlisting}
\vspace{-4pt}
{\small
% \halfspacing
\textcolor{violet}{Description:} sorts the elements in the list acording to their comparator. there are two reasons why lights should be resorted. first, if the lights have moved, that means their distance to the spatial changed. second, if the spatial itself moved, it means the distance from it to the individual lights might have changed. (FunCom)
\\ \\ %\vspace{-5pt}
%\hline
% 4
\singlespacing
}

\end{document}

\section{Dataset Details}

\textbf{\textsc{CodeSearchNet (CSN) Corpus}.} \citet{husain2019codesearchnet} released the \textsc{CSN Corpus}\footnote{\url{https://github.com/github/CodeSearchNet}}, a code search dataset for 6 programming languages.
Despite the authors' effort for data cleaning, in our observation, \textsc{CSN Corpus} is found to be one of the noisiest. The dataset contained duplicate descriptions, inseparable multi-words (e.g. \texttt{updateproductvariationlocalizedde- ltaprice, updatelocationinventory}), XML tags (e. g. \texttt{<tt>, <soup>, <sub>, <pre>, <p>}), non-English documentation, non-ASCII characters, escape characters, unwanted symbols (e.g. \texttt{@, \#, \{, \},} etc.), deprecated methods and descriptions, comments inside code, annotations (e.g. \texttt{@link, @code, @inheritdoc}) inside description, etc. Some datapoints (27,471) in our Python to Java translation are truncated by TransCoder, which have been marked with a special flag in our released dataset. 

%They collected a total of 6,452,226 functions in 6 programming languages: Go, Java, JavaScript, PHP, Python, Ruby. Among them, 2,326,976 functions contained natural language documentations.% We collected the Java and Python dataset from their data. Using a state-of-the-art program translation method, TransCoder \cite{transcoder}, we translate the Python source code to Java. %, and we collect the  1,046,493 functions with their corresponding documentation to our dataset. %Our program translation method is described in the next section. 

\textbf{DeepCom.} \citet{deepcom} released a dataset of 588,108 Java method and documentation pairs collected from 9,714 GitHub projects for code summarization\footnote{\url{https://github.com/xing-hu/DeepCom}}. Similar to \textsc{CodeSearchNet} \cite{husain2019codesearchnet}, they considered the first sentence of a documentation as the summary of the method as it typically describes the functionalities of Java methods. They filter out empty and single world descriptions and the setter, getter, constructors, and test methods, since they are easy for a model to summarize. In our manual analysis, we found HTML tags (e.g. \texttt{<tt>, <p>, <p class = …> , <li>, <ul>}), comment tags, annotations, escape characters inside descriptions, empty parentheses as descriptions, repetitive and non-meaningful descriptions, comments inside source code, etc. Despite the authors' claim, we found numerous test methods in the dataset, which were mostly meaningful data.

\textbf{CONCODE. } \citet{concode} released a dataset named CONCODE, collected by mining nearly 33,00 GitHub repositories\footnote{\url{https://github.com/sriniiyer/concode}}. In their preprocessed dataset, they replaced the names of the identifier and method names with generic terms, (e.g., \texttt{arg0, loc0, function,} etc.) and replaced all string literals with constants. This created a discrepancy with the other datasets, hence, we opted for their unprocessed dataset rather than the processed version. The unprocessed dataset released with CONCODE contained approximately 2.1 million Java methods and lowercased Javadoc-style document pairs. Upon duplicate removal, we were left with 733,878 datapoints. 

Although some noises were present in this dataset, we found this data to be least noisy in manual observation. We find that because of lower casing the documentations, some CamelCase tokens became inseparable. The dataset also contained non-English comments with English alphabets (mostly Italian). We found these documents hard to identify and remove.

\textbf{FunCom.} \citet{funcom} released a dataset of over 2.1 million pairs of Java methods and one-sentence method descriptions from over 28k Java projects\footnote{\url{http://leclair.tech/data/funcom/}}. They collected this dataset by filtering over 51 million Java methods from UCI Source Code datasets \citep{uci-data}. In their preprocessing step, \citet{funcom} removed all datapoints where the method is more than 100 tokens long, or the method description is over 13 tokens or below 3 tokens.

In our observation of this dataset, we found method descriptions containing HTML tokens (e.g. \texttt{<tt>}, annotations (e.g., \texttt{@link, @param}), comment tokens, unwanted symbols, solely non-alphabetic characters, etc. It also contained comments inside methods, and a large portion of the data were \texttt{getter}, \texttt{setter}, \texttt{tester}, and \texttt{toString} methods.

\section{Sample Data}

% \begin{figure*}
% 0
\begin{lstlisting}
Object getBean(String beanName){ 
	if(null == beanName){ 
		return null; 
	} 
	return applicationContext.getBean(beanName); 
}
\end{lstlisting}
\vspace{-4pt}
{\small
\halfspacing
\textcolor{violet}{Description:} this method is used to retrieve a bean by its name. note that this may result in new bean creation if the scope is set to ``prototype" in the bean configuration file. (CONCODE) \\ \vspace{-5pt}

% 3
\singlespacing
}
\vspace{-15pt}
\begin{lstlisting}
@Exported
    public boolean isIdle() {
        lock.readLock().lock();
        try {
            return workUnit == null && executable == null;
        } finally {
            lock.readLock().unlock();
        }
    }
\end{lstlisting}
\vspace{-4pt}
{\small
\halfspacing
\textcolor{violet}{Description:} returns true if this executor is ready for action. (CodeSearchNet) \\ \vspace{-5pt}

% 1
\singlespacing
}

\vspace{-15pt}
\begin{lstlisting}
public static void dbCommand(ParserArgs args){
  final Synergy synergy =(Synergy)args.get("synergy");
  if(synergy.reset){
    synergy.resetDb();
    synergy.update = true;
  }
  if(synergy.update){
    synergy.updateDb();
  }
}
\end{lstlisting}
\vspace{-4pt}
{\small
\halfspacing
\textcolor{violet}{Description:} manages synergy db state (CodeSearchNet- Python to Java) \\ \vspace{-5pt}
% 2
\singlespacing
}
\vspace{-15pt}
\begin{lstlisting}
public void sort (boolean transformChanged) { 
	if (list Size > 1){ 
		if (tlist == null || tlist.length != list.length){
			tlist = list.clone(); 
		} else { 
			System.arraycopy(list, 0, tlist, 0, list.length); 
		} 
		if (transform Changed) { 
    		for(int i = 0; i < listSize; i++) { 
    		    list[i]
    		    .computeLastDistance(owner); 
    		}
		} 
		SortUtil.msort(tlist, list, 0, list Size - 1, c); 
	} 
}
\end{lstlisting}
\vspace{-4pt}
{\small
\halfspacing
\textcolor{violet}{Description:} sorts the elements in the list acording to their comparator. there are two reasons why lights should be resorted. first, if the lights have moved, that means their distance to the spatial changed. second, if the spatial itself moved, it means the distance from it to the individual lights might have changed. (FunCom)
\\ \\ %\vspace{-5pt}
\hline

% 4
\singlespacing
}
\vspace{-15pt}
\begin{lstlisting}
protected void hideTabs(){
  if (getPageCount() <= 1) {
    setPageText(0,"");
    if (getContainer() instanceof CTabFolder) {
      ((CTabFolder)getContainer())
      .setTabHeight(1);
      Point point = getContainer().getSize();
      getContainer().setSize(point.x,point.y + 6);
    }
  }
}
\end{lstlisting}
\vspace{-4pt}
{\small
\halfspacing
\textcolor{violet}{Description:} if there is just one page in the multi - page editor part , this hides the single tab at the bottom. (DeepCom) \\ \vspace{-5pt}
\hline
% 5
\singlespacing
}

% \caption{Example code summary generated by Transformer model without and with CoDesc pretraining.}
\vspace{-14pt}
% \end{figure*}

\bibliographystyle{aclnatbib}
\bibliography{ref.bib}